\documentclass[]{article}

\usepackage{PRIMEarxiv}
\usepackage[utf8]{inputenc}
\usepackage[T1]{fontenc}
\usepackage{graphicx}
\usepackage{multirow}
\usepackage{amsmath,amssymb,amsfonts}
\usepackage{amsthm}
\usepackage{mathrsfs}
\usepackage{xcolor}
\usepackage{textcomp}
\usepackage{manyfoot}
\usepackage{booktabs}
\usepackage{listings}
\usepackage{macros}
\usepackage[round]{natbib}
\usepackage{comment}
\usepackage{enumitem}
\usepackage{blkarray}
\usepackage{bbm}
\pgfplotsset{compat=1.18}
\usepackage[hidelinks]{hyperref}    
\usepackage{url}          
\usepackage{nicefrac} 
\usepackage{microtype}
\usepackage{fancyhdr}
\usepackage{tikz}
\usepackage{tikzscale}
\usepackage{caption}
\usepackage{subcaption}
\usepackage{pgfplots}
\usepackage{xspace}
\usepackage{array}

\title{Online Dynamic Pricing of Complementary Products}

\author{
  Marco Mussi \\
  Politecnico di Milano \\
  \texttt{marco.mussi@polimi.it} \\
  \And
  Marcello Restelli \\
  Politecnico di Milano \\
  \texttt{marcello.restelli@polimi.it} \\
}

\allowdisplaybreaks[4]

\begin{document}

\maketitle

\begin{abstract}
Traditional pricing paradigms, once dominated by static models and rule-based heuristics, are increasingly being replaced by dynamic, data-driven approaches powered by machine learning algorithms. Despite their growing sophistication, most dynamic pricing algorithms focus on optimizing the price of each product independently, disregarding potential interactions among items. By neglecting these interdependencies in consumer demand across related goods, sellers may fail to capture the full potential of coordinated pricing strategies. In this paper, we address this problem by exploring dynamic pricing mechanisms designed explicitly for complementary products, aiming to exploit their joint demand structure to maximize overall revenue. We present an online learning algorithm considering both positive and negative interactions between products' demands. The algorithm utilizes transaction data to identify advantageous complementary relationships through an integer programming problem between different items, and then optimizes pricing strategies using data-driven and computationally efficient multi-armed bandit solutions based on heteroscedastic Gaussian processes. We validate our solution in a simulated environment, and we demonstrate that our solution improves the revenue w.r.t.~a comparable learning algorithm ignoring such interactions.
\end{abstract}

\section{Introduction}
\label{sec:intro}

The advent of digital technologies, combined with the chance to access large amounts of data, is reshaping the way businesses set prices, increasing the adoption of \emph{dynamic pricing} algorithms~\citep{den_boer_dynamic_2015}. These algorithms leverage advanced machine learning tools to analyze a multitude of factors, from consumer behavior and market trends to external variables such as product availability and competitor pricing. By adjusting prices in response to these dynamic factors, businesses employing dynamic pricing strategies aim to determine the optimal pricing strategy to maximize profit within a given time horizon. One of the key factors in dynamic pricing consists of understanding the \emph{demand} for a product. The demand represents the \emph{willingness} of the customers to buy an item, given a price. The extent to which demand varies as the price of an item changes defines the so-called \emph{elasticity} of a product. Such a demand usually takes into account \emph{only} the price at which we sell the item under analysis. However, considering all the products as independent prevents us from learning the correlations between the products in the catalog and other baskets, as well as carryover phenomena that can be exploited to maximize profit~\citep{yan_profit_2011}. The literature distinguishes between two types of product interdependencies: \emph{complementarity} and \emph{substitutability}. Complementary products \quotes{complement} each other~\citep{nicholson_microeconomic_2017}; these kinds of products are usually bought together (\eg a printer and its ink cartridges), and the sale of one implies an increment in the probability of selling the other. On the other hand, substitutable goods serve the same purpose and are rarely bought together; instead, they tend to cannibalize each other (\eg two identical items of different brands). Complementary and substitutable products have different effects on the demand function for the related products. The concept of \emph{cross-price elasticity} is useful to describe these effects: it measures the proportionate change in the quantity of a good demanded in response to a proportionate change in the price of some other good~\citep{nicholson_microeconomic_2017}. For complementary products, the cross-price elasticity is \emph{negative}: this is intuitive since a lower price of product $a$ leads to higher sales of product $a$ and, since the demand of product $a$ drives the demand of the complementary product $b$, we can also expect the quantity of product $b$ to increase. On the contrary, the effect for substitutable products is the opposite: the cross-price elasticity is \emph{positive} since the lower demand for product $c$ leads to an increase in the demand for its competing product $d$. The most challenging task in jointly optimizing pricing strategies involves identifying complementary relationships. Indeed, complementarity relationships exhibit characteristics such as \emph{asymmetry} and \emph{non-transitivity}~\citep{kocas_pricing_2018, yu_complementary_2019, xu_product_2020}. Furthermore, two products can be complementary while belonging to semantically different areas, so complementarity is inherently harder to capture and cannot be identified simply by similarity. On the other hand, substitutability relationships are symmetrical and transitive, and concern only products that satisfy the same need, which can be more easily identified also from a \emph{semantic} point of view.

\paragraph{Original Contributions.}
In this work, we aim to jointly optimize the pricing strategies of strictly related products to reach optimal solutions, taking into account the products' positive and negative interactions. We focus our attention on the more challenging problem of complementary relations and propose a computationally and statistically efficient solution to identify complementary products and determine their optimal pricing strategy, given a target index. Our solution avoids the request for information that is complex to retrieve and requires only purchase transaction data to detect complementary relations and estimate the optimal pricing strategy. More in detail, our contributions can be summarized as follows:
\begin{itemize}[leftmargin=16pt]
    \item In Section~\ref{sec:problemformulation}, we formulate the problem of finding the optimal pricing strategy in the presence of both positive and negative interactions among products. We introduce the setting, the decision variables, the underlying assumptions, and the corresponding learning problem. 
    \item In Section~\ref{sec:algorithm}, we introduce \algname (\algnameshort), a three-stage algorithm for optimizing pricing strategies in environments with interacting products. Sections~\ref{sec:algorithm:sub}, \ref{sec:algorithm:demand}, and \ref{sec:algorithm:compl} detail each stage of the algorithmic framework. We first aggregate substitutable products into meta-products, which can be efficiently identified and are mutually non-competing. Then, we design a new solution for detecting the most \emph{profitable} complementary relations, avoiding at the same time a combinatorial explosion of complexity \wrt the number of products in the catalog. Finally, we derive a computationally and statistically efficient procedure for learning the optimal pricing policies.    
    \item In Section~\ref{sec:experiments}, we validate our approach through a simulation study that allows us to test different levels of correlation among products. We compare our proposed methods against a baseline based on non-parametric regression models that price products independently. The results highlight the learning performance of our algorithms and provide insights into when exploiting inter-product relationships becomes particularly beneficial.
\end{itemize}
In Section~\ref{sec:background}, we present the relevant background needed to read the work properly. In Section~\ref{sec:related}, we discuss the relevant literature. Finally, in Section~\ref{sec:conclusions}, we summarize our work, and we draw some possible research perspectives.   

\paragraph{Notation.}
Throughout this paper, we interchangeably use lower- and upper-case letters (\eg $z, \ Z$) to indicate scalar values, lower-case boldface letters (\eg $\mathbf{v}$) to indicate (column) vectors, upper-case boldface letters (\eg $\Amx$) to indicate matrices, and calligraphic letters (\eg $\mathcal{X}$) to indicate sets. Given a vector or a matrix, we indicate with $(\cdot)^\top$ the transpose operator. Given a vector $\mathbf{v}$, we define $\mathsf{diag}(\mathbf{v})$ as the operator returning a diagonal matrix with $\mathbf{v}$ on the diagonal. Given a matrix $\Amx$, we indicate with $\determ(\Amx)$ its determinant. We indicate with $\mathbf{I}_Z$ the identity matrix of size $Z$ (we omit the index $Z$ when it can be inferred from the context). Given a scalar value $Z \in \mathbb{N}_{\geq 1}$, we define $\dsb{Z} \coloneqq \{ 1, 2, \ldots , Z \}$. Given a finite set $\Xs$, we denote with $|\Xs|$ its cardinality. Given an event/condition $e$, we define the indicator function $\indicator\{ e \}$, which takes the value $1$ if its argument $e$ is true, and $0$ otherwise. We use $\mathcal{O} ( \cdot )$ to present results -- on both statistical and computational complexity -- omitting constants.
\section{Gaussian Processes for Pricing}
\label{sec:background}

In this section, we provide the necessary background to fully understand this work.

Gaussian Processes~\citep[GPs,][]{rasmussen_gaussian_2006} are a powerful and flexible machine learning technique, widely used for \textit{regression} and \textit{classification} tasks. We focus on GPs for regression, where we want to learn a function $f : \Xs \to \mathbb{R}$, where $\Xs \in \mathbb{R}^d$ is the input space. To learn this function, we observe $T$ noisy samples of such a function. For every $t\in\dsb{T}$, samples are drawn from a Gaussian distribution with expected value $f(\xs)$, formally $y_t = f(\xs_t) + \epsilon_t$, where $\epsilon_t$ is a zero-mean $\sigma^2$-subgaussian random noise, independent conditioned to the past.\footnote{A zero-mean random variable $\xi$ is $\sigma^2$-subgaussian if $\mathbb{E}[\exp\left( \lambda \xi \right)] \le \exp\left(\frac{\lambda^2\sigma^2}{2}\right)$, for every $\lambda \in \mathbb{R}$.} A GP, denoted by $\mathcal{GP}(\mu(\cdot), k(\cdot, \cdot))$, is a collection of random variables $(f(\xs))_{\xs \in \Xs}$, such that every sub-collection of random variables $(f(\xs_i))_{i\in\dsb{J}}$ is jointly Gaussian with mean $\mathbb{E}[f(\xs_i)] = \mu(\xs_i)$, and covariance $\mathbb{E}[(f(\xs_i) - \mu(\xs_i))(f(\xs_j) - \mu(\xs_j))] = k(\xs_i, \xs_j)$.
We assume that function $f( \cdot )$ belongs to a \emph{reproducing kernel Hilbert space} (RKHS) $\mathcal{H}_k$~\citep{berlinet2004reproducing}.

Given the vector of points $[\xs_1 , \ldots , \xs_T]^\top$, we define the \emph{kernel matrix} $\Ks_T = [k(\xs_i, \xs_j)]_{i,j \in \dsb{T}}$. Then, considering the corresponding vector of observed rewards $\ys_{T} = [y_1, \ldots, y_T]^\top$, and given a target point $\xs$, we can compute a GP estimate of the model $\widehat{\mu}_T(\xs)$ and the uncertainly $\widehat{\sigma}_T^2(\xs)$:
\begin{align*}
\widehat{\mu}_T(\xs) &= \ks_T(\xs)^\top \ (\Ks_T + \sigma^2 \mathbf{I}_T)^{-1} \ \ys_T \\
\widehat{\sigma}_T^2(\xs) &= k(\xs, \xs) - \ks_T(\xs)^\top \ (\Ks_T + \sigma^2 \mathbf{I}_T)^{-1} \ \ks_T(\xs) 
\end{align*}
where $\ks_T(\xs) = [k(\xs_1, \xs), \ldots, k(\xs_T, \xs)]^\top$.\footnote{This expression of $\widehat{\sigma}_T^2(\xs)$ considers just the epistemic uncertainty of the model, not the aleatoric one.}

The kernel should be chosen according to the covariance model over the different samples in the problem under analysis. Due to its versatility, the \emph{squared exponential} (SE) kernel is the most commonly used. It is defined as: 
\begin{align*}
k(\xs, \xs') = \exp\left(-\frac{\|\xs - \xs'\|^2_2}{2\ell^2}\right),
\end{align*}
where $\ell > 0$ represents the \emph{length scale}. The kernel's hyperparameters (in this case, just $\ell$) are typically optimized by maximizing the marginal likelihood.

\subsection{Online Learning with Gaussian Processes}
\label{sec:background:onlinealearning}
GPs can be used to learn \emph{online} the best point $\xs^* \in \argmax_{\xs \in \Xs} f(\xs)$, \ie the one maximizing function $f(\cdot)$. Several algorithms allow us to find $\xs^*$ minimizing, at the same time, the cumulative regret $\widetilde{R}(\bm{\pi}, T) \coloneqq T f(\xs^*) - \sum_{t = 1}^T f(\xs_t)$, where samples $\xs_t$ are drawn according to algorithm/policy $\bm{\pi}$.\footnote{This objective is equivalent to maximizing the cumulative reward during the interaction.}
Nearly-optimal algorithms~\citep{scarlett2017lower} in this sense are \texttt{GP-UCB}~\citep{srinivas2010gaussian} and its improvement \texttt{IGP-UCB}~\citep{chowdhury_kernelized_2017}. In \texttt{IGP-UCB}, which we will adopt later in this paper, at every time step $t\in\dsb{T}$ we want to play an optimistic action, according to the usual exploration/exploitation trade-off we have in online learning. To select such an action, the posterior mean $\widehat{\mu}_{t-1}(\xs)$ and uncertainty $\widehat{\sigma}_{t-1}^2(\xs)$ are computed using all samples available up to time $t-1$, for every $\xs \in \Xs$. The best action is then selected according to:
\begin{align}\label{eq:opt_choice}
    \xs_t \in \argmax_{\xs \in \Xs} \ \widehat{\mu}_{t-1}(\xs) + \beta_t \sqrt{\widehat{\sigma}^2_{t-1}(\xs)},
\end{align}
with:
\begin{align}\label{eq:opt_bound}
    \beta_t = B + \sqrt{2 \sigma^2 (\gamma_{t-1} + 1 + \lognat(1/\delta))},
\end{align}
where $B$ is an upper bound on the RKHS norm ($\|f\|_k \leq B$) and $\gamma_{t-1} \coloneqq \lognat ( \determ (\mathbf{I}_{t-1} + \sigma^{-2}\Ks_{t-1})) / 2$ is the maximum information gain at time $t-1$, that for SE kernels is $\mathcal{O}((\lognat t)^{d+1})$. For more details, we refer the reader to~\citep{srinivas2010gaussian,chowdhury_kernelized_2017,vakili2021information}.

\paragraph{Gaussian Processes for Bernoulli Random Variables.}
GPs are naturally designed to work with Gaussian random variables. However, in pricing, we aim to model demand curves, which are designed by learning a demand model starting from Bernoulli samples. Nevertheless, the versatility of GPs allows them to extend beyond purely Gaussian settings and to model a broader class of subgaussian random variables~\citep{chowdhury_kernelized_2017}. This fact enables the chance to use them to model Bernoulli random variables, as every random variable bounded in the range $[a,b]$ is $\sigma^2$-subgaussian with $\sigma^2 = (b-a)^2/4$. Given this, a Bernoulli random variable is subgaussian with $\sigma^2 = 1/4$.\footnote{At the moment, there are no specific results on GPs with Bernoulli rewards~\citep{mussi2024open}.}

\paragraph{Computational Complexity.}
Computing $\widehat{\mu}_{T}$ and $\widehat{\sigma}^2_{T}$ requires the inversion of kernel matrix $\Ks_T$, which scales as $\mathcal{O}(T^3)$ with the number of training points $T$. If we want to utilize online learning solutions in this setting, we need to perform such estimates iteratively, which leads to an overall complexity of $\mathcal{O}(T^4)$. Such computational burden, even if polynomial \wrt $T$, is very demanding from the practical perspective. It can be reduced by observing that kernel matrices are incrementally built by adding one row and one column to the previous kernel matrix, and thus enables the chance to use \emph{block-wise matrix} inversion~\citep{lu2002inverses} and reduce the overall complexity of online learning with GPs to $\mathcal{O}(T^3)$. 
\section{Problem Formulation}
\label{sec:problemformulation}

In this section, we formalize the problem we address in this paper. We start in Section~\ref{sec:problemformulation:setting} by defining the setting and the related assumptions. Then, in Section~\ref{sec:problemformulation:learningproblem}, we present the learning problem, and we define the metric we use for evaluating the quality of our solution.

\subsection{Setting}
\label{sec:problemformulation:setting}
We want to find the optimal pricing strategy for a set of products $\mathcal{P}$, with $|\mathcal{P}|=P$. Our goal is, given a time horizon $T$, to set for every time $t \in \dsb{T}$ a vector of percentage margins (from now on, margins) $\ms_t = (m_{1,t}, \, \ldots, \, m_{P,t})$ where $m_{i,t} \in \mathcal{M}$ is the price we choose for product $i$ at time $t$, and $\mathcal{M} \subset \mathbb{R}$ is the set of possible margins, which we assume to be finite with $|\mathcal{M}|=M$. We can change our pricing strategy at every time step $t \in \dsb{T}$. A single time step $t$ can be an arbitrary, given, time interval (\eg a week). We define the (percentage) margin $m_{i,t}$ as:
\begin{equation*}
    m_{i,t} \coloneqq \frac{p_{i,t} - c_{i}}{c_{i}},
\end{equation*}
where $p_{i,t}$ is the selling price for product $i$ at time $t$, and $c_{i}$ its acquisition cost. For a generic product $i \in \mathcal{P}$, we denote as $\overline{v}_{i}(\ms)$ the average volumes (sales), which we assume to depend on the whole margin vector $\ms$. We assume that, given the number of impressions $n_{i,t}$ (\ie the number of times the product $i$ and its price is shown to a customer at time $t$) of a product and its demand function $d_i(\ms_t)$, the number of realized sales are modeled as a binomial random variable with mean $d_i(\ms_t)$ and number of trials $n_{i,t}$, that is: $v_{i,t} \sim \mathrm{Binomial}\left(n_{i,t},\, d_i(\ms_t)\right)$, so that the expected value satisfies $\mathbb{E}[v_{i,t}(\ms_t)] = \overline{v}_i(\ms_t) = \overline{n}_{i} \, d_i(\ms_t)$, where $\overline{n}_{i}$ is the average number of impressions for product $i$ in a single time step.

We consider a scenario in which we have both positive and negative interactions among the products, \ie we assume that the purchase of a product may affect the purchase probability of all the other products. We identify two types of relationships between products: \emph{substitutability} and \emph{complementarity}. Consider two products $a, b \in \mathcal{P}$. On the one hand, we call $a$ and $b$ substitutable products if an increase in the sales of one product implies a decrease in the sales of the other. On the other hand, we call $a$ and $b$ complementary products if an increase in the sales of one product implies an increase in the sales of the other.

\paragraph{Assumptions and Available Data.}
We consider non-perishable products with unlimited availability. These assumptions, nowadays, hold in several cases. For example, the unlimited availability virtually holds for e-commerce websites adopting the \emph{dropshipping}~\citep{gurpreet2018dropshipping} paradigm. We assume to be in a stationary environment, where both demand curves and acquisition costs remain constant over time.\footnote{The extension to non-stationary environments has been discussed several times for the independent product pricing~\citep{bauer_optimal_2018,nambiar_dynamic_2019, javanmard_multi-product_2020,mussi_pricing_2022} and is out of the scope of this work.} 
We consider a scenario in which we do not have access to information related to the products we are selling to be more general, except for the information related to substitutable goods, \ie the one satisfying the same need.\footnote{As already discussed in Section~\ref{sec:intro}, this is a mild request as these relations are easy to detect.} We only make the mild assumption to have access to transaction data reporting all the sales for every product $i \in \mathcal{P}$, divided by baskets, and provided with the timestamp of the sale. We also assume that we can observe when a customer chooses not to buy a product to be able to compute demand curves.\footnote{This is a mild assumption as well in practice, as, for example, an e-commerce website can keep track of the times in which a webpage of an item has been visited.}

\subsection{Learning Problem}
\label{sec:problemformulation:learningproblem}
The goal of our learning problem is to find the vector of the optimal margin $\ms^*$, \ie the vector maximizing our objective function $f(\ms)$. Formally:
\begin{equation}
\ms^* \in \argmax_{\ms \in \mathcal{M}^P} f(\ms), 
\end{equation}
where the objective function $f(\ms)$ is the \emph{profit}:
\begin{equation}
    f(\ms) \coloneqq \sum_{i \in \dsb{P}} f_{i}(\ms) = \sum_{i \in \dsb{P}} m_{i} \ c_{i} \ \overline{v}_{i}(\ms),
    \label{eq:total_profit}
\end{equation}
over all the products.\footnote{The objective function can be made more general by considering a convex combination of revenue and profit. In this case, the objective function $f$ becomes $f(\ms) \coloneqq \sum_{i \in \dsb{P}} ( m_{i} + \alpha ) \ c_{i} \ v_{i}(\ms)$, with $\alpha \in [0,1]$. In this formulation, it is easy to observe that if we select $\alpha = 0$ we optimize the \emph{profit} (as in Equation~\ref{eq:total_profit}), while if we select $\alpha = 1$ we optimize the \emph{revenue}.} We call $\bm{\pi} = (\pi_t)_{t\in\dsb{T}}$ the (history-dependent) policy returning at each time $t$ a vector of margins $\ms_t$. We define the cumulative reward of such a policy as:
\begin{equation}
R(\bm{\pi}, T) \coloneqq \sum_{t \in\dsb{T}} f(\ms_t).
\end{equation}
The goal of our algorithm is to find a policy maximizing the expected cumulative reward $ \mathbb{E} \left[ R(\bm{\pi}, T) \right] $, where the expectation is taken over the randomness of the realizations and the possible randomness of the algorithm.\footnote{This is equivalent to minimizing the expected cumulative regret $\, \mathbb{E} [ \widetilde{R}(\bm{\pi}, T) ] \coloneqq T f(\ms^*) - \mathbb{E} [R(\bm{\pi}, T)].$}
\section{Algorithm: Overview}
\label{sec:algorithm}

\begin{figure}[t!]
   \centering
   \resizebox{\linewidth}{!}{\input{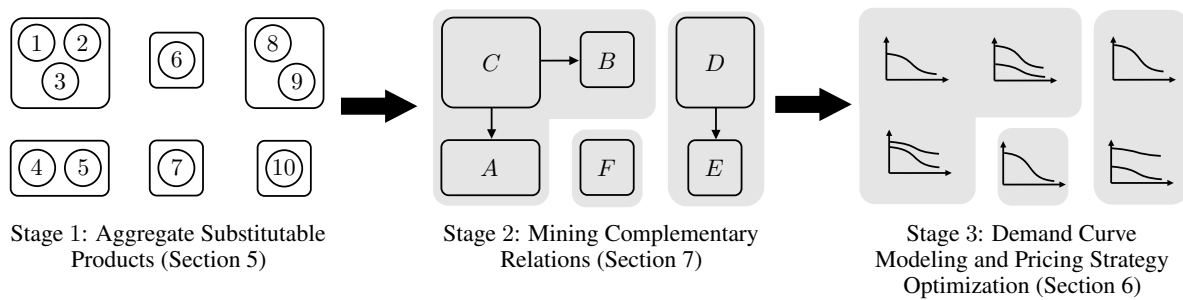}}
   \caption{Algorithm outline.}
   \label{fig:algorithm_outline}
\end{figure}

In this section, we present an overview of \algname (\algnameshort), an algorithm that identifies and optimizes the pricing strategy of complementary products in an online fashion.
The algorithm takes as input the product catalog, including information about substitutable products, and all the order records, and provides as output a coherent pricing strategy for all products. The algorithm is structured in $3$ stages, which can be summarized as follows:
\begin{enumerate}[leftmargin=50pt,topsep=2pt,itemsep=2pt,label={(Stage \arabic*)}]
    \item \textit{Aggregate Substitutable Products}: all the products satisfying the same need are aggregated in meta-products. These new meta-products will be the actual products we use in the next stages (we will refer to them simply as products for the sake of simplicity, as the original products are no longer considered). This operation is performed only once at the algorithm's startup and is repeated only in the case of a change in the catalog.
    \item \textit{Mining Complementary Relations}: all the products are analyzed to search for the \textit{leader-follower sets} which are more profitable to exploit. This operation is repeated at regular intervals, as we want to avoid performing it at every time step to strike a balance between the algorithm's performance and computational complexity.
    \item \textit{Demand Curve Modeling and Pricing Strategy Optimization}: We utilize previously detected complementary relations to estimate demand curves, considering also \textit{cross-selling} dynamics. These demand curves are then used to define an exploratory pricing strategy.
\end{enumerate}
A graphical representation of such a pipeline is provided in Figure~\ref{fig:algorithm_outline}, where an example of this aggregation is illustrated. We can observe how $1$, $2$, and $3$ satisfy the same need, and they are aggregated into the meta-product $C$. In the next phases, we consider only meta-products, such as $C$. Then, we identify complementary relations and define the leaders and followers. In the figure, we observe how $C$ and $D$ are leaders, $A$, $B$, and $E$ are followers, while $F$ is not related to any other item. Finally, we define the demand curve for these clusters of products and their subsequent pricing strategies.

In the following sections, we present the three stages in detail. We first present the stage in which we aggregate substitutable products and the rationale behind this operation (Section~\ref{sec:algorithm:sub}), then we present our demand learning model for both independent and complementary products (Section~\ref{sec:algorithm:demand}), and, finally we present our solution to detect the relevant complementary relations (Section~\ref{sec:algorithm:compl}). The rationale behind this inverted order \wrt the algorithm's flow is that to understand the solution to detect complementary products, we need to know how to estimate demand curves.

\section{Algorithm: Aggregate Substitutable Products}
\label{sec:algorithm:sub}
In this section, we discuss why we need to aggregate substitutable products. Indeed, the first step of our algorithm involves aggregating substitutable products into meta-products, each representing a specific \emph{need}. Then, we will define (Sections~\ref{sec:algorithm:demand} and~\ref{sec:algorithm:compl}) the optimal margin for the meta-product that will be applied to all the products that constitute it.\footnote{More elaborate solutions on how we can find margins for products that are part of a meta-product should be considered. However, we believe this problem deserves a standalone work.} 

The need to cluster substitutable products stems from the characteristics of these types of products. 
Two substitutable products compete with each other and are often subject to the phenomenon of cannibalization~\citep{moorthy_market_1992} since they satisfy the same need. Indeed, pricing these products separately would exacerbate cannibalization and could lead to them competing against each other, resulting in a reduction of profit. Clustering these products together and considering them as a single product, where the objective function is the total sales, removes the chance that they will cannibalize each other. This problem of cannibalization can be viewed from a game-theoretic perspective, where we can choose whether to opt for a cooperative solution or two competitive algorithms when pricing two substitutable products. We know that the optimal \emph{stable} solution for the cooperative case is always better or equal than the one for non-cooperative games (\ie a possibly suboptimal \emph{Nash equilibrium},~\citealt{nash1950equilibrium,nash1951non}). In the remainder of this paper, we refer to meta-products calling them products for the sake of simplicity, as the original products are no longer considered.

\section{Algorithm: Demand Curves Modeling and Pricing Strategy Optimization}
\label{sec:algorithm:demand}

In this section, we introduce an efficient approach to demand learning based on heteroscedastic Gaussian processes. Although in general the demand for a product can depend on the prices of all other products, modeling such dependencies is highly challenging due to the extremely large action space. Consequently, it becomes essential to find a way to reduce both the statistical complexity (i.e., the number of samples needed for learning) and the computational complexity (i.e., the number of operations required to estimate the optimal margin or action). To address this, we adopt a \emph{leader-follower} framework, wherein the set of products is partitioned into disjoint subsets, each consisting of one \emph{leader} and zero, one, or more \emph{followers}. When a leader has no followers, we say the product is \emph{independent} of the others. In this section, we assume that the leader-follower sets are given; the procedure for determining these associations is deferred to Section~\ref{sec:algorithm:compl}. Our objective is to construct the vector of margins $\ms_t \in \mathcal{M}^P$ for each time step $t \in \dsb{T}$. Given the partition into leader-follower sets, we compute the optimal margins by processing each set individually. We begin by presenting the demand model and the resulting strategy for selecting the optimistic margin for \emph{independent products} (Section~\ref{sec:algorithm:demand:independent}), and then we extend our analysis to the case of \emph{leader-follower sets} (Section~\ref{sec:algorithm:demand:leaderfollower}).

\subsection{Independent Products}
\label{sec:algorithm:demand:independent}

Given an independent product $i$, we now want, assuming that its demand depends only on its own margin, to estimate such a demand from historical data and perform an optimistic exploration to learn its optimal margin $m_{i}^*$. 

What we observe during a single time step (\eg a week) is a sequence of visits to, \eg the online page of a product, and then we observe the realization of the Bernoulli indicating whether the customer bought or not the product. In principle, we can create a demand curve in this way using standard GPs between margins and a binary-valued variable to represent the realization of a Bernoulli ($1$ if the product has been sold, $0$ otherwise), generating in this way a model of the demand curve, considering that every sample has $\sigma^2 = 1/4$.\footnote{This is an upper bound to the variance of the Bernoulli, which corresponds to the real variance only if $\mathbb{E}[y_{i,t}|m_{i,t}] = 1/2$, the real variance may be lower but we cannot do better than upper bound it.} 
However, this approach incurs a significant computational burden at each time step $t \in \dsb{T}$, with a per-step complexity of $\mathcal{O}(t^3 \overline{n}_i^3)$. This results in an overall complexity of $\mathcal{O}(T^4 \overline{n}_i^3)$.\footnote{While this can be mitigated to some extent through incremental matrix inversion~\citep{lu2002inverses}, the computational cost remains prohibitively high.} Afford such complexity is not feasible, especially in the context of the pipeline we aim to implement. In the following, we present how heteroscedastic GPs can be leveraged to substantially reduce this computational overhead.

\paragraph{Demand Learning via Heteroscedastic GPs --- A First Solution.}
We can aggregate information by creating a data structure in which we store information about sales. Such a structure at time $t$ for independent products should contain, for every $\tau\in\dsb{t}$ the margin $m_{i,\tau}$ for product $i$ at time $\tau$, its actual sales $v_{i,\tau}$ and the number $n_{i,\tau}$ of \emph{impressions}, \ie the times the product and the related margin has been observed (no matter if a customer bought it or not) during the time step $\tau$. With this information, we can estimate the demand in a computationally efficient way using heteroscedastic GPs. In this model, given a target margin $m$, at time $t+1$ we can compute:
\begin{align}
\widehat{\mu}_{i,t}(m) &= \ks_{i,t}(m)^\top \ \left( \Ks_{i,t} + \mathsf{diag}\left(\frac{\sigma^2}{\ns_{i,t}}\right)\right)^{-1} \ \ys_{i,t} \label{eq:independent:muhat} \\
\widehat{\sigma}_{i,t}^2(m) &= k(m, m) - \ks_{i,t}(m)^\top \ \left( \Ks_{i,t} + \mathsf{diag}\left(\frac{\sigma^2}{\ns_{i,t}}\right)\right)^{-1} \ \ks_{i,t}(m) \label{eq:independent:sigmahat}
\end{align}
with kernel matrix $\Ks_{i,t}$ and vectors $\ns_{i,t}$, $\ys_{i,t}$ and $\ks_{i,t}$ are defined as:
\begin{align*}
& \Ks_{i,t} \coloneqq [k(m_{i,j},m_{i,h})]_{j,h\in\dsb{t}} \\
& \ns_{i,t} \coloneqq [n_{i,1}, \ldots, n_{i,t}]^\top \\
& \ys_{i,t} \coloneqq [y_{i,1}, \ldots, y_{i,t}]^\top \\
& \ks_{i,t}(m) \coloneqq [k(m_{i,1}, m), \ldots, k(m_{i,t}, m)]^\top 
\end{align*}
where $y_{i,\tau} = v_{i,\tau} / n_{i,\tau}$, \ie the number of sales divided by the number of impressions for product $i$ at time $\tau$ (this value represents a realization of the demand). The rationale behind this model is that the quality of every sample $y_{i,\tau}$ depends on the number of impressions $n_{i,\tau}$, \ie the number of times we observed the realization of the Bernoulli for such a margin. Indeed, we can add $n_{i,\tau}$ samples whose average is $y_{i,\tau}$ with variance $\sigma^2$ each or add just one sample with variance $\sigma^2 / n_{i,\tau}$, we get the same result~\citep{binois2018practical}. In this way, the computational complexity of learning online in this setting can drop to $\mathcal{O}(T^3)$ (if we consider efficient matrix inversion updates). 

\paragraph{Demand Learning via Heteroscedastic GPs --- A More Efficient Solution.}
Even if the solution above improves the algorithm's efficiency, the computational complexity is still too high. We can improve it by using the same idea of considering an equivalent GP with heteroscedastic noise, but considering one sample for each margin only. To do so, we have to build a dataset in which we have sales statistics $\widetilde{v}_{i,t,j}$ and $\widetilde{n}_{i,t,j}$ for product $i$ at time $t$ for each margin $m_j$ with $j\in\dsb{M}$. Such counters are defined as: 
\begin{align*}
    & \widetilde{v}_{i,t,j} = \!\! \sum_{\tau\in\dsb{t}} \! v_{i,\tau} \cdot \indicator\{ m_{i,\tau} = m_j \}  \ \qquad \widetilde{n}_{i,t,j} = \!\! \sum_{\tau\in\dsb{t}} \! n_{i,\tau} \cdot \indicator\{ m_{i,\tau} = m_j \} \ \qquad \widetilde{y}_{i,t,j} = \frac{\widetilde{v}_{i,t,j}}{\widetilde{n}_{i,t,j}}.
\end{align*}
Given such aggregated counters, we can design a new heteroscedastic GP (whose kernel matrix size will be bounded by the number of actions $M$) as:
\begin{align}
\widehat{\mu}_{i,t}(m) &= \widetilde{\ks}_{i,t}(m)^\top \ \left( \widetilde{\Ks}_{i,t} + \mathsf{diag}\left(\frac{\sigma^2}{\widetilde{\ns}_{i,t}}\right)\right)^{-1} \widetilde{\ys}_{i,t} \label{eq:independent:muhat:margins} \\
\widehat{\sigma}_{i,t}^2(m) &= k(m, m) - \widetilde{\ks}_{i,t}(m)^\top \ \left( \widetilde{\Ks}_{i,t} + \mathsf{diag}\left(\frac{\sigma^2}{ \widetilde{\ns}_{i,t}}\right)\right)^{-1} \ \widetilde{\ks}_{i,t}(m) \label{eq:independent:sigmahat:margins}
\end{align}
where matrix $\widetilde{\Ks}_{i,t} \coloneqq [k(m_{j},m_{h})]_{j,h\in\dsb{M}}$ and vectors $\widetilde{\ns}_{i,t}$, $\widetilde{\ys}_{i,t}$ and $\widetilde{\ks}_{i,t}$ are defined as:
\begin{align*}
& \widetilde{\ns}_{i,t} \coloneqq [\widetilde{n}_{i,t,1}, \ldots, \widetilde{n}_{i,t,M}]^\top \\
& \widetilde{\ys}_{i,t} \coloneqq [\widetilde{y}_{i,t,1}, \ldots, \widetilde{y}_{i,t,M}]^\top \\
& \widetilde{\ks}_{i,t}(m) \coloneqq [k(m_1, m), \ldots, k(m_M, m)]^\top.
\end{align*}
We highlight that the heteroscedastic GP of Equations~\eqref{eq:independent:muhat} and~\eqref{eq:independent:sigmahat} and the one of Equations~\eqref{eq:independent:muhat:margins} and~\eqref{eq:independent:sigmahat:margins} gives the same result in terms of $\widehat{\mu}_{i,t}(m)$ and $\widehat{\sigma}_{i,t}^2(m)$, for every $m\in\mathcal{M}$~\citep{binois2018practical}.\footnote{In the model above, margins $m_j \in\mathcal{M}$ whose corresponding $n_{i,t,j}$ is $0$ should not be included in the GP model, as their variance will be undefined. In this case, no efficient matrix update is possible.}

\paragraph{Optimistic Action Choice.}
Once at time $t$ we have an estimate of $\widehat{\mu}_{i,t-1}(m)$ and $\widehat{\sigma}_{i,t-1}^2(m)$ for every $m\in\mathcal{M}$, we can make use of the optimistic bound of \igpucb~\citep{chowdhury_kernelized_2017} to optimistically select the margin maximizing our objective function:
\begin{align*}
    m_{i,t} \in \argmax_{m\in\mathcal{M}} \ \widehat{f}_{i,t}(m),
\end{align*}
where $\widehat{f}_{i,t}(m)$ is the optimistic counterpart of $f_{i}(\ms)$ presented in Equation~\eqref{eq:total_profit}:
\begin{align}\label{eq:independent:opt_choice}
    \widehat{f}_{i,t}(m) = m \ c_i \; \widehat{d}^{\text{\texttt{ OPT}}}_{i,t-1}(m) \; \widehat{n}_{i,t-1},
\end{align}
where $\widehat{n}_{i,t-1}$ is an estimate of the average number of impressions per-stage calculated with data up to time $t-1$, and the estimated optimistic demand $\widehat{d}^{\text{\texttt{ OPT}}}_{i,t-1}(m)$ is calculated as: 
\begin{align}
\widehat{d}^{\text{\texttt{ OPT}}}_{i,t-1}(m) =  \widehat{\mu}_{i,t-1}(m) + \beta_t \sqrt{\widehat{\sigma}^2_{i,t-1}(m)} , 
\end{align}
with $\beta_t$ chosen as in Equation~\eqref{eq:opt_bound}, considering $\sigma^2=1/4$, as we need to consider the variance proxy of the original process, and using as maximum information gain~\citep{binois2018practical}:
$$\gamma_{i,t-1} = \frac{1}{2} \lognat(\determ(\mathbf{I} + \mathbf{D}_{i,t-1}\widetilde{\Ks}_{i,t-1} \mathbf{D}_{i,t-1})),$$
where $\mathbf{D}_{i,t-1} = \mathsf{diag}\left(\sigma^2 / \widetilde{\ns}_{i,t-1}\right)^{-1/2}$.

In this optimistic bound, the exploration term is applied just to the estimate of the demand curve, as we have no statistical uncertainty on terms $m$ and $c_i$. Regarding $\widehat{n}_{i,t-1}$, the reason for which we do not need an exploration bonus is twofold. First, this term is just a rescaling factor and does not change the optimal margin we select. Second, we do not need any form of exploration on that, given that we will refine its estimate at any step, no matter the chosen margin. 

\paragraph{Computational Complexity.}
The regression model presented above significantly mitigates the computational complexity, as learning online requires $\mathcal{O}(TM^3)$ operations, where $M$ is the number of margins, which means a strong improvement \wrt $\mathcal{O}(T^3)$ given that we usually have $M \ll T$.

\subsection{Leader-Follower Sets}
\label{sec:algorithm:demand:leaderfollower}

In this part, we consider the pricing of a single set of complementary products. Then, we will repeat the process for all the other sets. In this set $\mathcal{S}$, we assume to have one leader and one or more followers. With a little abuse of notation, we call the leader $l$ and the followers $f_i$, for $i\in\dsb{F}$ where $F$ is the number of followers of the set under analysis. We have to compute, given a vector of margins $(m_l, m_{f_1}, \ldots , m_{f_F})$, the overall estimated return from all the products belonging to this set. This is given by:
\begin{align}\label{eq:set:opt_choice}
    \widehat{f}_{\mathcal{S},t}(m_l, m_{f_1}, \ldots , m_{f_F}) = \widehat{f}_{l,t}(m_l) + \sum_{i\in\dsb{F}} \widehat{f}_{f_i,t}(m_l,m_{f_i}),
\end{align}
and our goal is to choose:
\begin{align}\label{eq:argmax_over_set}
    (m_{l,t}, m_{f_1,t}, \ldots , m_{f_F,t}) \in \argmax_{(m_l, m_{f_1}, \ldots , m_{f_F})\in\mathcal{M}^{F+1}} \ \widehat{f}_{\mathcal{S},t}(m_l, m_{f_1}, \ldots , m_{f_F}).
\end{align}
This means we want to jointly optimize the prices we choose for the leader and the followers. In this formulation, the demand of the leader depends only on the margin of the leader $m_l$, while the demand of a follower $f_i$ depends on the margin of such a follower $m_{f_i}$ and the one of its leader $m_l$. As such, the optimistic objective function of the leader $\widehat{f}_{l,t}(m_l)$ is defined as:
\begin{align*}
    \widehat{f}_{l,t}(m_l) = c_l \; m_l \; \widehat{d}^{\text{\texttt{ OPT}}}_{l,t-1}(m_l)  \; \widehat{n}_{l,t-1} ,
\end{align*}
where $\widehat{n}_{l,t-1}$ is the average number of impressions we observed (per stage) up to stage $t-1$ for leader product $l$, and the optimistic demand is:
\begin{align*}
     \widehat{d}^{\text{\texttt{ OPT}}}_{l,t-1}(m_l) = \widehat{\mu}_{l,t-1}(m_l) + \beta_t \sqrt{\widehat{\sigma}^2_{l,t-1}(m_l)}.
\end{align*}
On the other hand, $\widehat{f}_{f_i,t}(m_l,m_{f_i})$ is defined as:
\begin{align*}
    \widehat{f}_{f_i,t}(m_l,m_{f_i}) = c_{f_i} \; m_{f_i} \; \widehat{d}^{\text{\texttt{ OPT}}}_{f_i,t-1} \; \widehat{n}_{f_i,t-1},
\end{align*}
where $\widehat{d}^{\text{\texttt{ OPT}}}_{f_i,t-1}$ is the equivalent optimistic demand for the follower $f_i$, defined as: 
\begin{align*}
    \widehat{d}^{\text{\texttt{ OPT}}}_{f_i,t-1} = & \ \ \widehat{p}_{l,f_i,t-1} \, \widehat{\mu}_{l,t-1}(m_l) \; \widehat{\overline{d}}^{\text{\texttt{ OPT}}}_{f_i,t-1} + \left( (1-\widehat{p}_{l,f_i,t-1}) + \widehat{p}_{l,f_i,t-1} (1 - \widehat{\mu}_{l,t-1}(m_l)) \right) \; \widehat{\widetilde{d}}^{\text{\texttt{ OPT}}}_{f_i,t-1},
\end{align*}
where $\widehat{p}_{l,f_i,t-1}$ is the sample-based estimate up to time $t-1$ of $\, \overline{p}_{l,f_i}$, \ie the probability that a user observes $f_i$ when leader $l$ is in the basket, and the demands are defined as: 
\begin{align*}
    \widehat{\overline{d}}^{\text{\texttt{ OPT}}}_{f_i,t-1} = \ \widehat{\overline{\mu}}_{f_i,t-1}(m_{f_i}) + \beta_t \sqrt{\widehat{\overline{\sigma}}^2_{f_i,t-1}(m_{f_i})} \qquad \text{and} \qquad \widehat{\widetilde{d}}^{\text{\texttt{ OPT}}}_{f_i,t-1} = \ \widehat{\widetilde{\mu}}_{f_i,t-1}(m_{f_i}) + \beta_t \sqrt{\widehat{\widetilde{\sigma}}^2_{f_i,t-1}(m_{f_i})},
\end{align*}
where $\widehat{\overline{\mu}}_{f_i,t-1}(m_{f_i})$ and $\widehat{\overline{\sigma}}^2_{f_i,t-1}(m_{f_i})$ are the estimators related to the demand computed using only the impressions and sales generated for product $f_i$ when leader $l$ is in the same basket, and $\widehat{\widetilde{\mu}}_{f_i,t-1}(m_{f_i})$ and $\widehat{\widetilde{\sigma}}^2_{f_i,t-1}(m_{f_i})$ the ones related demand computed using the sales and impressions generated when product $l$ is not in the same basket. The rationale is to trade off the sales we can expect for the follower, taking into account both the probability that a customer concurrently sees the leader and the probability that such a customer buys the leader product. 
All the demands are estimated using the model proposed in Section~\ref{sec:algorithm:demand:independent}, taking care of creating a new dataset in which we divide the impressions, and the sales generated must be stored together with the information about the products bought together. 

\paragraph{Computational Complexity.}
These regression models present the same computational complexity order per-product as that of Section~\ref{sec:algorithm:demand:independent}, \ie $\mathcal{O}(TM^3)$. However, once we have computed all models for all products in the cluster, we also need to search for the optimal margin vector. To maintain an acceptable level of computational complexity, the search for the optimal solution (Equation~\ref{eq:argmax_over_set}) within the given space must be carefully structured. This necessity arises because the search space lacks convexity/concavity properties, which preclude the use of gradient-based optimization methods and instead roughly require a grid-search approach, leading to a complexity in the order of $\mathcal{O}(M^{F+1})$.\footnote{The grid search can be substituted with more elaborated heuristic approaches~\citep[see, \eg][]{vikhar2016evolutionary}.} If we accept that all followers share the same margin (\ie $m_{f_1,t} = m_{f_2,t} = \dots = m_{f_F,t}$), the complexity of this search is limited to $\mathcal{O}(M^2)$ for each set (this will introduce a suboptimality in the maximum revenue reachable).

\section{Algorithm: Mining Complementarity Relations}
\label{sec:algorithm:compl} 

In this section, we focus on identifying complementary relations among the products we aim to price. These relations, in general, may involve numerous products with intricate interaction dynamics, posing significant challenges for effective modeling. Detecting general complementary relations requires navigating an extensive hypothesis space, which can lead to issues such as model variance and sensitivity to noise. Moreover, our goal extends beyond merely identifying products that are frequently purchased together. Instead, we aim to discover products whose optimal joint pricing maximizes overall value, surpassing the benefits of pricing them independently. To mitigate such a high complexity, we focus specifically on \emph{leader-follower} complementary relations. In this framework, each product is classified as either ($i$) a leader to one or more products, ($ii$) a follower of another product, or ($iii$) not involved in any complementary relation (independent).\footnote{For computational efficiency, a product designated as a follower cannot simultaneously act as a leader for other products, as this would significantly increase the complexity of determining the optimal prices, given that all the generated subsets will not be independent and we cannot perform an independent optimization of the different subsets.}

We begin by presenting the concept intuitively, providing a foundation for understanding. Then, we introduce a practical example that helps transition to a formal definition of the problem. 

The goal is to determine the optimal partitioning of a set of $P$ products to maximize revenue. To accomplish this, using the models introduced in Section~\ref{sec:algorithm:demand}, we construct a real-valued $P \times P$ matrix $\mathbf{V} = [v_{i,j}]_{i,j\in\dsb{P}}$ that represents the optimistic revenues for all possible leader-follower pairs. In this matrix, on-diagonal elements $v_{i,i}$ represent the highest possible revenue obtainable when product $i$ is priced independently (see Section~\ref{sec:algorithm:demand:independent}), \ie $v_{i,i} = \max_{m\in\mathcal{M}} \widehat{f}_{i,t}(m)$ (Equation~\ref{eq:independent:opt_choice}). On the other hand, off-diagonal elements $v_{i,j}$ (with $i \neq j$) in the matrix corresponds to the maximum estimated revenue from treating product $i$ as the leader and product $j$ as the follower, with their margins jointly optimized (see Section~\ref{sec:algorithm:demand:leaderfollower}) to maximize the objective function (Equation~\ref{eq:set:opt_choice}) minus the value of the leader $v_{i,i}$ if priced independently (to avoid to count it several times if it is leader of more than one follower).\footnote{From the practical perspective, it may be useful to subtract a small penalty factor to off-diagonal elements to avoid the detection of \emph{ghost} relations, as the joint performance is lower bounded by the independent performance, and the noise may easily lead to detect false complementary relations.}

With this matrix in place, we want to identify the most profitable combinations of products. With this aim, we define a second $P \times P$ binary-valued matrix $\mathbf{X}$. Here, each element $x_{i,j}$ indicates whether product $i$ is selected as the leader of product $j$ ($x_{i,j} = 1$) or not ($x_{i,j} = 0$). To determine the optimal solution, we formulate a \emph{binary programming} problem that maximizes the sum of all the elements of the matrix resulting from the Hadamard (element-wise) product between the matrices $\mathbf{V}$ and $\mathbf{X}$, subject to a set of constraints governing the placement of zeros and ones in $\mathbf{X}$ we will present later.

To better understand this solution, before formalizing the problem, we present an example of what we expect as output. 

\noindent\textbf{Example.}~~\textit{Consider a set of $P=6$ products. Suppose we compute, using our models, the expected return of all the couples, and we get a matrix $\mathbf{V}$ as follows:}
\begin{align}
\begin{blockarray}{ccccccc}
  & A & B & C & D & E & F \\
\begin{block}{c[cccccc]}
A & 10 & 20 & 30 & 40 & 50 & 60 \\
B & 10 & 20 & 30 & 40 & 50 & 60 \\
C & \blue{\bm{30}} & \blue{\bm{50}} & \blue{\bm{30}} & 40 & 50 & 60 \\
D & 10 & 20 & 30 & \blue{\bm{40}} & \blue{\bm{80}} & 60 \\
E & 10 & 20 & 30 & 40 & 50 & 60 \\
F & 10 & 20 & 30 & 40 & 50 & \blue{\bm{60}} \\
\end{block}
\end{blockarray}   \label{eq:example:values}     
\end{align}
\textit{One optimal solution involves selecting $C$ as the leader for products $A$ and $B$, selecting $D$ as the leader for $E$, and leaving $F$ independent. The corresponding binary matrix $\mathbf{X}$ is:}
\begin{align}
\begin{blockarray}{ccccccc}
  & A & B & C & D & E & F \\
\begin{block}{c[cccccc]}
A & 0 & 0 & 0 & 0 & 0 & 0 \\
B & 0 & 0 & 0 & 0 & 0 & 0 \\
C & \blue{\bm{1}} & \blue{\bm{1}} & \blue{\bm{1}} & 0 & 0 & 0 \\
D & 0 & 0 & 0 & \blue{\bm{1}} & \blue{\bm{1}} & 0 \\
E & 0 & 0 & 0 & 0 & 0 & 0 \\
F & 0 & 0 & 0 & 0 & 0 & \blue{\bm{1}} \\
\end{block}
\end{blockarray} \label{eq:example:binary}       
\end{align}
\textit{In this illustrative example, it is easy to observe that the estimated reward is given by the sum of the values of the matrix in Equation~\eqref{eq:example:values} in the positions where in Equation~\eqref{eq:example:binary} we have the terms $1$. Given that, the overall estimated return can be seen as the sum of all the elements in the matrix $\mathbf{V} \circ \mathbf{X}$, where $\circ$ is the Hadamard product.} \hfill$\square$

After presenting the example, we formalize the problem as an integer programming problem with linear constraints~\citep{wolsey2020integer}. The objective function aims to maximize the sum of all elements in the matrix obtained from the element-wise product:
\begin{align*}
    \sum_{i \in \dsb{P}} \sum_{j \in \dsb{P}} x_{i,j} \; v_{i,j},
\end{align*}
where $v_{i,j}$ are positive real-valued elements, and $x_{i,j}$ are the binary variables over which we optimize. If product $i$ is a leader (\ie $i\in\mathcal{L}$), the following conditions (constraints) must hold:
\begin{align}
    & \textstyle \sum_{j \in \dsb{P} \setminus \{i\}} x_{i,j} \geq 1, \qquad \ \, \forall i \in \mathcal{L} \label{eq:optproblem:constraint:11} \\
    & \textstyle \sum_{j \in \dsb{P} \setminus \{i\}} x_{j,i} = 0, \qquad \ \: \forall i \in \mathcal{L} \label{eq:optproblem:constraint:12} \\
    & \ \textstyle x_{i,i} = 1, \qquad\qquad\qquad\quad \ \, \forall i \in \mathcal{L} \label{eq:optproblem:constraint:13}
\end{align}
where Equation~\eqref{eq:optproblem:constraint:11} ensures that the leader has at least one follower, Equation~\eqref{eq:optproblem:constraint:12} imposes that it has no leaders (it must not be a follower of any other product), and Equation~\eqref{eq:optproblem:constraint:13} ensures we consider the value of the leader product in the formulation. 

On the other hand, if product $i$ is a follower (\ie $i\in\mathcal{F}$) or independent (\ie $i\in\mathcal{I}$), it has to satisfy the following conditions:
\begin{align}
    & \textstyle \sum_{j \in \dsb{P}} x_{j,i} = 1, \qquad\quad \ \; \: \forall i \in \mathcal{F} \cup \mathcal{I} \label{eq:optproblem:constraint:21} \\
    & \textstyle \sum_{j \in \dsb{P} \setminus \{i\}} x_{i,j} = 0, \qquad \, \forall i \in \mathcal{F} \cup \mathcal{I} \label{eq:optproblem:constraint:22}
\end{align}
where Equation~\eqref{eq:optproblem:constraint:21} ensures that each follower is linked to exactly one product (including itself if independent), and Equation~\eqref{eq:optproblem:constraint:22} ensures that a follower does not lead any other product.

To enforce that each product is \emph{either} a leader or a follower/independent, we impose the following condition:
\begin{equation}
    ( \text{Eq.}~\eqref{eq:optproblem:constraint:11} \ \land \ \text{Eq.}~\eqref{eq:optproblem:constraint:12} \ \land \ \text{Eq.}~\eqref{eq:optproblem:constraint:13}  ) \ \oplus \ ( \text{Eq.}~\eqref{eq:optproblem:constraint:21} \ \land \ \text{Eq.}~\eqref{eq:optproblem:constraint:22} )
\end{equation}
where $\land$ and $\oplus$ represent the logical \emph{and} and the \emph{exclusive or}, respectively. This will ensure that $\mathcal{L} \cup \mathcal{F} \cup \mathcal{I} = \mathcal{P}$ and their (even pair-wise) intersections are the empty set. To implement the \emph{exclusive or}, we consider $P$ additional binary optimization variables $(z_i)_{i\in\dsb{P}}$ into our integer programming problem. As an implementation choice, a variable $z_i$ will be equal to $0$ if the corresponding product is a leader and $1$ otherwise. With these additional quantities, we can define our optimization problem by implementing the \emph{exclusive or} using linear constraints. The integer programming problem is then defined as:
\begin{align}
    \max_{\stackrel{(x_{i,j})_{i,j\in\dsb{P}}}{(z_i)_{i\in\dsb{P}}}} & \ \ \sum_{i \in \dsb{P}} \sum_{j \in \dsb{P}} x_{i,j} \; v_{i,j} \\
    \text{subject to:} \quad 
    & \textstyle 0 \leq x_{i,j} \leq 1, \qquad\qquad\qquad\qquad\quad \ \ \: \forall i, j \in \dsb{P} \label{eq:optproblem:constraint:bin_x} \\
    & \textstyle 0 \leq z_i \leq 1, \qquad\qquad\qquad\qquad\qquad \ \, \forall i \in \dsb{P} \label{eq:optproblem:constraint:bin_y} \\
    & \textstyle \sum_{j \in \dsb{P} \setminus \{i\}} x_{i,j} \geq 1 - P z_i, \qquad\quad \ \ \; \! \forall i \in \dsb{P} \label{eq:optproblem:constraint:11a} \\
    & \textstyle \sum_{j \in \dsb{P} \setminus \{i\}} x_{j,i} \leq 0 + P z_i, \qquad\quad \ \ \, \forall i \in \dsb{P} \label{eq:optproblem:constraint:12a} \\
    & \textstyle x_{i,i} \leq 1 + P z_i, \qquad\qquad\qquad\qquad\quad \! \forall i \in \dsb{P} \label{eq:optproblem:constraint:13a} \\
    & \textstyle x_{i,i} \geq 1 - P z_i, \qquad\qquad\qquad\qquad\quad \! \forall i \in \dsb{P} \label{eq:optproblem:constraint:13b} \\
    & \textstyle \sum_{j \in \dsb{P}} x_{j,i} \geq 1 - P (1 - z_i), \qquad \ \ \, \forall i \in \dsb{P} \label{eq:optproblem:constraint:21a} \\
    & \textstyle \sum_{j \in \dsb{P}} x_{j,i} \leq 1 + P (1 - z_i), \qquad \ \ \, \forall i \in \dsb{P} \label{eq:optproblem:constraint:21b} \\
    & \textstyle \sum_{j \in \dsb{P} \setminus \{i\}} x_{i,j} \leq 0 + P (1 - z_i), \quad \;\! \forall i \in \dsb{P} \label{eq:optproblem:constraint:22a}
\end{align}
Some comments are in order. First, we observe that now the optimization variables are $P^2+P$, even if the ones involved in the objective function still remain $P^2$. Second, we use $z_i$ variables to determine the type of product and the constraints to be applied. In Constraints~\eqref{eq:optproblem:constraint:11a}, \eqref{eq:optproblem:constraint:12a}, \eqref{eq:optproblem:constraint:13a} and~\eqref{eq:optproblem:constraint:13b} referring to leader products, we add/subtract (given the type of constraint) a term $P \, z_i$ to \emph{lift} the constraints in the case in which the product under analysis is a follower/independent.\footnote{For simplicity and readability, terms $P \, z_i$ and $P \, (1-z_i)$ are used consistently, although smaller values could suffice to lift the constraints in some cases.} For equality constraints $x = Z$, we impose these constraints by imposing that $x \leq Z$ and $x \geq Z$.\footnote{In some cases, we just need one constraint as the other is already satisfied by construction.} In Constraints~\eqref{eq:optproblem:constraint:21a}, \eqref{eq:optproblem:constraint:21b} and~\eqref{eq:optproblem:constraint:22a}, we do the opposite, and to model the case in which products are followers/independent, and the term $P \, (1-z_i)$ lift the constraints in the case we are facing a leader.

It is important to note that when there is more than one follower for a single leader, this optimization problem yields an \emph{optimistic estimate} of the impact of joint product pricing. In fact, given a leader $l$, the margin $m_l$ that maximizes the overall revenue for a follower $f_i$ may be suboptimal for another follower $f_j$. Addressing this issue rigorously would significantly increase the computational complexity. Therefore, we accept this slight suboptimality to maintain a reasonable computational burden.

\paragraph{Computational Complexity.}
Finding the optimal solution for binary programming problems is known to be \textsf{NP}-hard in general~\citep{karp1972reducibility,schrijver1998theory,papadimitriou1998combinatorial,wolsey1999integer}. Given that, we should make considerations and adopt strategies for reducing the computational burden. First, we can determine a rough split of the products, generating disjoint subsets. We should then run the algorithm multiple times, once on each subset, which significantly reduces the computational complexity of this pipeline step. Moreover, this step should not be performed at every $t\in\dsb{T}$. The reason for avoiding running this procedure at every step is twofold. First, to reduce computational complexity, as this operation is computationally heavy. Second, to prevent frequent changes to product pairings during learning, as this could negatively impact pricing stability (see Section~\ref{sec:experiments}). 
\section{Experimental Validation}
\label{sec:experiments}

In this section, we empirically assess the effectiveness of the proposed \algname (\algnameshort) algorithm through a set of controlled simulations. The goal of this experimental campaign is twofold. First, we aim to evaluate the capability of \algnameshort to learn optimal pricing strategies in environments with different types of product interactions. Second, we seek to compare its performance with suitable baselines to isolate the contribution of each algorithmic component. We start by introducing the algorithms under comparison, including two \algnameshort variants and a baseline which disregards product interdependencies. Then, we design two distinct experimental scenarios. First, in Section~\ref{sec:experiments:indep}, we consider the case in which product demands are independent. Then, in Section~\ref{sec:experiments:compl}, we consider the case where products exhibit complementary relations which may influence the optimal behavior. Finally, in Section~\ref{sec:experiments:computationalperf}, we analyze the computational performance of all algorithms to assess their scalability and practical viability. The code used in this section is available at \url{https://github.com/marcomussi/ComplementaryProductsPricing}.

\paragraph{Evaluated Algorithms and Baseline.}
We consider two versions of the \algnameshort algorithm and a coherent baseline. The first variant, \algknowngraph (\emph{known graph}), considers an implementation of \algnameshort in which we assume that the complementarity graph between products is known in advance. In this version, the algorithm focuses exclusively on learning demand curves and optimizing prices given a fixed interaction structure. This configuration allows us to isolate and assess the effectiveness of the demand modeling and pricing optimization stages in a simpler scenario where the relational structure does not need to be inferred. The second variant, \algunknowngraph (\emph{unknown graph}), represents the complete version of the algorithm. In this version, neither the structure of the complementarity graph nor the demand functions are known. The algorithm must therefore learn the underlying relations between the products by solving the integer programming problem described in Section~\ref{sec:algorithm:compl}, while simultaneously learning the demand models and updating pricing strategies online. This setup enables the assessment of the algorithm’s capacity to autonomously discover profitable complementary relations and exploit them for coordinated pricing. We run the \algunknowngraph by recomputing the graph through the integer programming problem at every step, even though it can be run at regular intervals to reduce computational complexity. To provide a meaningful point of comparison, we also include an independent pricing baseline, denoted as \indepalg (\texttt{Independent Products Pricing}). This algorithm employs the same learning framework as \algnameshort for modeling demand and optimizing prices, but treats each product as independent, ignoring any potential cross-product relations (following the model presented in Section~\ref{sec:algorithm:demand:independent}). By comparing \algknowngraph and \algunknowngraph against \indepalg, we can quantify the advantage of explicitly modeling interactions among complementary products and evaluate the impact of structure learning on both performance and computational cost.

\subsection{Independent Environment}
\label{sec:experiments:indep}

We start by analyzing the learning performances of the algorithms in the case of independent demand curves.

\paragraph{Setting.}
We consider scenarios involving different sets of products $\mathcal{P}$ with varying cardinalities $P \in \{5, 10, 20\}$. For each product, we define a discrete set of possible margins $\mathcal{M} = \{0.1, 0.3, 0.5, 0.7, 0.9\}$, which can be applied to all the products. At each round, we simulate $100$ impressions per product, assuming a concurrence probability $\overline{p}_{i,j}=1$ for every $i,j\in\mathcal{P}$ for the sake of simplicity in the presentation of the results. Demand curves are synthetically generated under the constraint of being monotonically decreasing with respect to the margin, reflecting standard economic assumptions (this assumption is not needed for the correct execution of the algorithms). The algorithms' performance is evaluated in terms of the obtained rewards. Specifically, we report the instantaneous rewards (mean $\pm$ $95\%$ confidence interval), while the optimum is represented as a horizontal line in the figures.\footnote{We report the instantaneous reward instead of the cumulative one, as in this way it is simpler to visualize when we reach the optimum.} The optimal value is computed via Monte Carlo simulations using $10^5$ samples for each action to ensure accurate estimation.

\begin{figure}[t!]
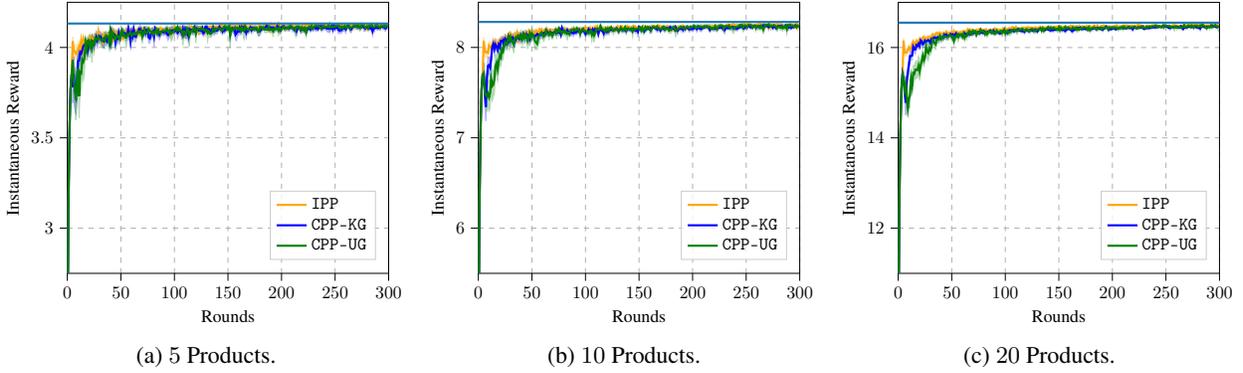

    \centering
    \begin{subfigure}[b]{0.331\textwidth}
        \resizebox{\linewidth}{!}{\input{img/exp/IndependentEnv_T300_prods5_trials30}}
        \caption{$5$ Products.}
        \label{fig:exp:indep:5prods}
    \end{subfigure}
    \hfill
    \begin{subfigure}[b]{0.32\textwidth}
        \resizebox{\linewidth}{!}{\input{img/exp/IndependentEnv_T300_prods10_trials30}}
        \caption{$10$ Products.}
        \label{fig:exp:indep:10prods}
    \end{subfigure}
    \hfill
    \begin{subfigure}[b]{0.327\textwidth}
        \resizebox{\linewidth}{!}{\input{img/exp/IndependentEnv_T300_prods20_trials30}}
        \caption{$20$ Products.}
        \label{fig:exp:indep:20prods}
    \end{subfigure}
    \caption{Independent demands environments ($30$ trials, mean $\pm$ $95\%$ C.I.).}
    \label{fig:exp:indep}
\end{figure}

\paragraph{Results.}
The results are presented in Figure~\ref{fig:exp:indep}.
In this scenario, regardless of the number of products considered, all algorithms achieve optimal performance, confirming that the demand learning model behaves as expected. In particular, the independent pricing algorithm (\indepalg) exhibits faster convergence to the optimal reward. This outcome is reasonable, as \indepalg represents the simplest model among the three, yet it retains sufficient representational capacity to learn the optimal action correctly in this simpler scenario. This behavior becomes even more evident as the number of products increases. In Figure~\ref{fig:exp:indep:10prods}, which considers $10$ products, and in Figure~\ref{fig:exp:indep:20prods}, which extends the analysis to $20$ products, we observe that both competing algorithms are outperformed by \indepalg in the first few rounds of the learning process. In particular, \algunknowngraph shows, as expected, the worst performance at the beginning as it has to explore more to learn also the graph structure.

\subsection{Complementary Environments}
\label{sec:experiments:compl}

We now move to the scenario in which we consider product interactions influencing demands. 

\paragraph{Setting.}
We consider a setting in which product interactions arise between demand curves following a leader–follower structure. As in the previous experiment, we analyze scenarios involving different sets of products $\mathcal{P}$ with cardinalities $P \in \{5, 10, 20\}$. For each product, we define a discrete set of possible margins $\mathcal{M} = \{0.1, 0.3, 0.5, 0.7, 0.9\}$, which can be applied uniformly across all products. At each round of the algorithm, we simulate $100$ impressions per product, assuming a concurrence probability $\overline{p}_{i,j}=1$ for every $i,j \in \mathcal{P}$. When a follower product is proposed after a customer purchases its leader, we model the increase in demand as a multiplicative factor applied to the base demand (i.e., the demand the product would exhibit in the absence of the leader). Two levels of demand interaction are considered, corresponding to increases of $10\%$ and $40\%$, to emulate two levels of interaction. As in the previous experiment, the performance of the proposed methods is evaluated in terms of instantaneous rewards, and the optimal value is estimated via Monte Carlo simulations with $10^5$ samples per action.

\begin{figure}[t!]
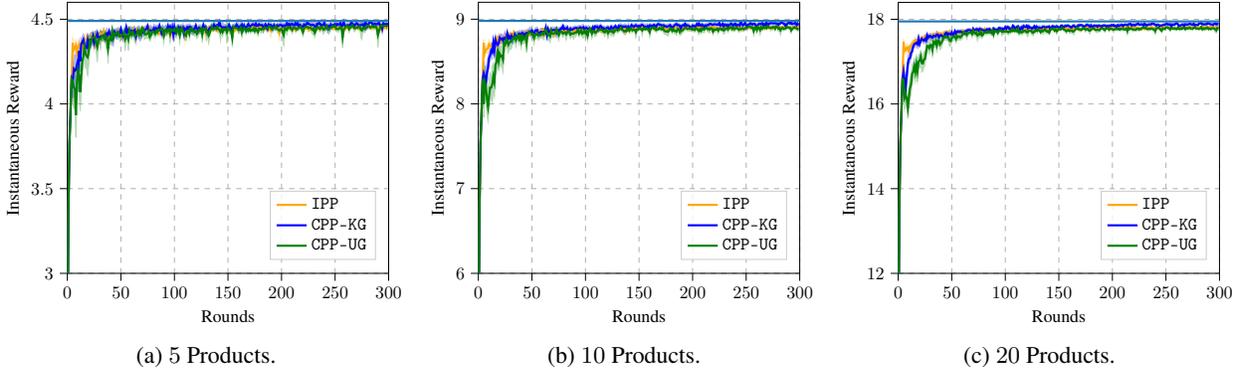

    \centering
    \begin{subfigure}[b]{0.331\textwidth}
        \resizebox{\linewidth}{!}{\input{img/exp/SimpleComplementaryEnv_T300_prods5_trials30}}
        \caption{$5$ Products.}
        \label{fig:exp:compl_low:5prods}
    \end{subfigure}
    \hfill
    \begin{subfigure}[b]{0.32\textwidth}
        \resizebox{\linewidth}{!}{\input{img/exp/SimpleComplementaryEnv_T300_prods10_trials30}}
        \caption{$10$ Products.}
        \label{fig:exp:compl_low:10prods}
    \end{subfigure}
    \hfill
    \begin{subfigure}[b]{0.327\textwidth}
        \resizebox{\linewidth}{!}{\input{img/exp/SimpleComplementaryEnv_T300_prods20_trials30}}
        \caption{$20$ Products.}
        \label{fig:exp:compl_low:20prods}
    \end{subfigure}
    \caption{Complementary demand environments with mild complementarity dynamics ($30$ trials, mean $\pm$ $95\%$ C.I.).}
    \label{fig:exp:compl_low}
\end{figure}

\begin{figure}[t!]
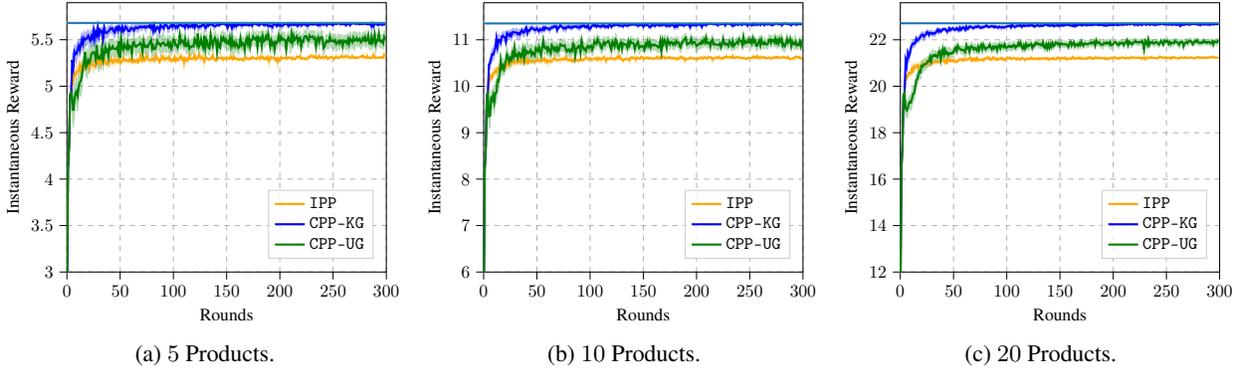

    \centering
    \begin{subfigure}[b]{0.329\textwidth}
        \resizebox{\linewidth}{!}{\input{img/exp/ComplementaryEnv_T300_prods5_trials30}}
        \caption{$5$ Products.}
        \label{fig:exp:compl:5prods}
    \end{subfigure}
    \hfill
    \begin{subfigure}[b]{0.325\textwidth}
        \resizebox{\linewidth}{!}{\input{img/exp/ComplementaryEnv_T300_prods10_trials30}}
        \caption{$10$ Products.}
        \label{fig:exp:compl:10prods}
    \end{subfigure}
    \hfill
    \begin{subfigure}[b]{0.325\textwidth}
        \resizebox{\linewidth}{!}{\input{img/exp/ComplementaryEnv_T300_prods20_trials30}}
        \caption{$20$ Products.}
        \label{fig:exp:compl:20prods}
    \end{subfigure}
    \caption{Complementary demands environments ($30$ trials, mean $\pm$ $95\%$ C.I.).}
    \label{fig:exp:compl}
\end{figure}

\paragraph{Results.}
In Figures~\ref{fig:exp:compl_low} and~\ref{fig:exp:compl}, we show the results for the complementary environment for the demand increases of $10\%$ and $40\%$, respectively.
We can observe from Figure~\ref{fig:exp:compl_low} that, across all product sets, when the increment in the follower’s demand is mild ($10\%$), all algorithms exhibit comparable performance and successfully reach the optimal reward. The same considerations discussed in the previous case of independent items also apply here, in terms of convergence speed and behavior, as the number of products increases. In contrast, Figure~\ref{fig:exp:compl}, where the follower’s demand increases significantly ($40\%$), illustrates a markedly different pattern. In this setting, where the pull effect becomes substantial, the independent algorithm converges rapidly but only to a suboptimal reward, while the algorithms exploiting product complementarities achieve higher performance. In particular, the algorithm that does not need to learn the complementary graph structure reaches the optimum, whereas the version that simultaneously learns this structure also converges, but at a slower rate. A detailed analysis suggests that this slower convergence is due to an over-exploration required for learning the graph of complementarity relationships. The two figures thus reveal distinct behaviors, raising the question of why, in the mild increment scenario, the algorithms that exploit complementarities do not show any advantage. We observed that the reason lies in the nature of the optimal actions. When the increment is small, the optimal solution obtained for each product individually coincides with that obtained by jointly optimizing over the cluster of complementary items. Conversely, in the high-increment scenario, the optimal joint action for the leader–follower group differs from the one derived by treating products independently. In particular, we observe that reducing the margin of the leader decreases its individual reward, but simultaneously induces a boost in the follower’s demand, ultimately leading to a higher overall reward for the cluster.

\subsection{Computational Performances}
\label{sec:experiments:computationalperf}

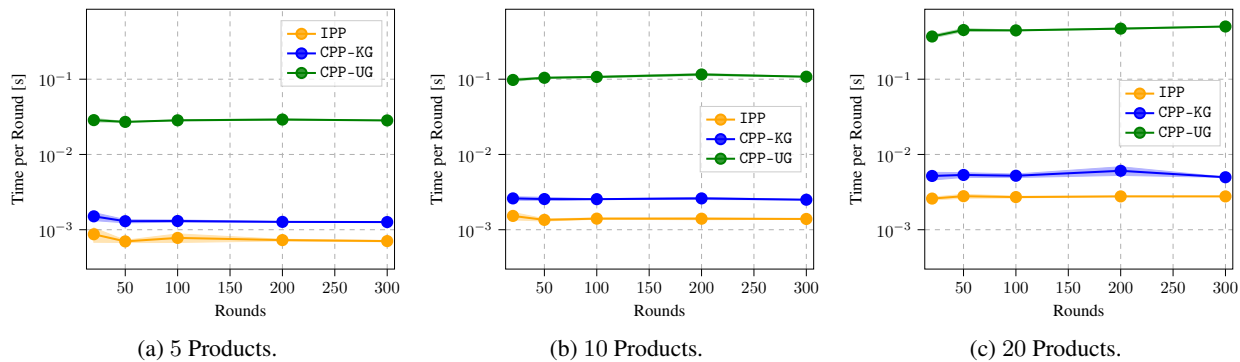
\begin{figure}[t!]
    \centering
    \begin{subfigure}[b]{0.325\textwidth}
        \resizebox{\linewidth}{!}{% This file was created with tikzplotlib v0.10.1.
\begin{tikzpicture}

\definecolor{darkgray176}{RGB}{176,176,176}
\definecolor{green}{RGB}{0,128,0}
\definecolor{lightgray204}{RGB}{204,204,204}
\definecolor{orange}{RGB}{255,165,0}

\begin{axis}[
height=7cm,
width=8cm,
legend cell align={left},
legend style={
  fill opacity=0.8,
  draw opacity=1,
  text opacity=1,
  at={(0.96,0.83)},
  anchor=east,
  draw=lightgray204
},
log basis y={10},
tick align=outside,
tick pos=left,
xlabel={Rounds},
ylabel={Time per Round [s]},
x grid style={darkgray176},
xmin=13, xmax=307,
xtick style={color=black},
y grid style={darkgray176},
ymin=0.0003, ymax=0.85,
ymode=log,
ytick style={color=black},
grid=both,
grid style={dashed,gray!50!white},
ytick={1e-05,0.0001,0.001,0.01,0.1,1},
yticklabels={
  \(\displaystyle {10^{-5}}\),
  \(\displaystyle {10^{-4}}\),
  \(\displaystyle {10^{-3}}\),
  \(\displaystyle {10^{-2}}\),
  \(\displaystyle {10^{-1}}\),
  \(\displaystyle {10^{0}}\)
}
]
\path [draw=orange, fill=orange, opacity=0.3]
(axis cs:20,0.0010589528707172)
--(axis cs:20,0.000681409773478361)
--(axis cs:50,0.000675433499384163)
--(axis cs:100,0.000677096200301558)
--(axis cs:200,0.000709872088113493)
--(axis cs:300,0.000691441802149101)
--(axis cs:300,0.000721197021042223)
--(axis cs:300,0.000721197021042223)
--(axis cs:200,0.000746413150152498)
--(axis cs:100,0.000880494761154741)
--(axis cs:50,0.000723861353826286)
--(axis cs:20,0.0010589528707172)
--cycle;

\path [draw=blue, fill=blue, opacity=0.3]
(axis cs:20,0.00169145250127706)
--(axis cs:20,0.00132245636179057)
--(axis cs:50,0.0012338335897769)
--(axis cs:100,0.00126114962875169)
--(axis cs:200,0.00126052508629915)
--(axis cs:300,0.00125259524073844)
--(axis cs:300,0.00127598733219858)
--(axis cs:300,0.00127598733219858)
--(axis cs:200,0.00128217472754363)
--(axis cs:100,0.00134908463180739)
--(axis cs:50,0.00136254473663912)
--(axis cs:20,0.00169145250127706)
--cycle;

\path [draw=green, fill=green, opacity=0.3]
(axis cs:20,0.0302375172784946)
--(axis cs:20,0.0269473267385342)
--(axis cs:50,0.0268649294756119)
--(axis cs:100,0.0280324400087955)
--(axis cs:200,0.028509708488551)
--(axis cs:300,0.0281664828128015)
--(axis cs:300,0.0284796301060523)
--(axis cs:300,0.0284796301060523)
--(axis cs:200,0.0296938356513111)
--(axis cs:100,0.0288037955144284)
--(axis cs:50,0.0273111331083114)
--(axis cs:20,0.0302375172784946)
--cycle;

\addplot [very thick, orange, mark=*, mark size=3, mark options={solid}]
table {%
20 0.000870181322097778
50 0.000699647426605225
100 0.000778795480728149
200 0.000728142619132996
300 0.000706319411595662
};
\addlegendentry{\indepalg}
\addplot [very thick, blue, mark=*, mark size=3, mark options={solid}]
table {%
20 0.00150695443153381
50 0.00129818916320801
100 0.00130511713027954
200 0.00127134990692139
300 0.00126429128646851
};
\addlegendentry{\algknowngraph}
\addplot [very thick, green, mark=*, mark size=3, mark options={solid}]
table {%
20 0.0285924220085144
50 0.0270880312919617
100 0.0284181177616119
200 0.029101772069931
300 0.0283230564594269
};
\addlegendentry{\algunknowngraph}
\end{axis}

\end{tikzpicture}}
        \caption{$5$ Products.}
        \label{fig:exp:runningtime:5prods}
    \end{subfigure}
    \hfill
    \begin{subfigure}[b]{0.325\textwidth}
        \resizebox{\linewidth}{!}{% This file was created with tikzplotlib v0.10.1.
\begin{tikzpicture}

\definecolor{darkgray176}{RGB}{176,176,176}
\definecolor{green}{RGB}{0,128,0}
\definecolor{lightgray204}{RGB}{204,204,204}
\definecolor{orange}{RGB}{255,165,0}

\begin{axis}[
height=7cm,
width=8cm,
legend cell align={left},
legend style={
  fill opacity=0.8,
  draw opacity=1,
  text opacity=1,
  at={(0.96,0.5)},
  anchor=east,
  draw=lightgray204
},
log basis y={10},
tick align=outside,
tick pos=left,
xlabel={Rounds},
ylabel={Time per Round [s]},
x grid style={darkgray176},
xmin=13, xmax=307,
xtick style={color=black},
y grid style={darkgray176},
ymin=0.0003, ymax=0.85,
ymode=log,
ytick style={color=black},
grid=both,
minor y tick num=9,
minor grid style={dotted,gray!40!white},
grid style={dashed,gray!50!white},
ytick={1e-05,0.0001,0.001,0.01,0.1,1},
yticklabels={
  \(\displaystyle {10^{-5}}\),
  \(\displaystyle {10^{-4}}\),
  \(\displaystyle {10^{-3}}\),
  \(\displaystyle {10^{-2}}\),
  \(\displaystyle {10^{-1}}\),
  \(\displaystyle {10^{0}}\)
}
]
\path [draw=orange, fill=orange, opacity=0.3]
(axis cs:20,0.00170095053565116)
--(axis cs:20,0.00134940775978952)
--(axis cs:50,0.00130637830531219)
--(axis cs:100,0.00137715224133973)
--(axis cs:200,0.00135798351673457)
--(axis cs:300,0.00137960048984906)
--(axis cs:300,0.00140157194463987)
--(axis cs:300,0.00140157194463987)
--(axis cs:200,0.00144427569003728)
--(axis cs:100,0.00143147536409851)
--(axis cs:50,0.00139644342625519)
--(axis cs:20,0.00170095053565116)
--cycle;

\path [draw=blue, fill=blue, opacity=0.3]
(axis cs:20,0.00274746736183126)
--(axis cs:20,0.00246787229881327)
--(axis cs:50,0.00241563283787343)
--(axis cs:100,0.00252094322914281)
--(axis cs:200,0.0025204019209755)
--(axis cs:300,0.0024913320165984)
--(axis cs:300,0.00251882731694855)
--(axis cs:300,0.00251882731694855)
--(axis cs:200,0.00269371369868393)
--(axis cs:100,0.00257450526481471)
--(axis cs:50,0.00269520223750499)
--(axis cs:20,0.00274746736183126)
--cycle;

\path [draw=green, fill=green, opacity=0.3]
(axis cs:20,0.102396440475025)
--(axis cs:20,0.093085019619427)
--(axis cs:50,0.102927517766958)
--(axis cs:100,0.1051656626649)
--(axis cs:200,0.111774908777548)
--(axis cs:300,0.107296179064937)
--(axis cs:300,0.108533515047205)
--(axis cs:300,0.108533515047205)
--(axis cs:200,0.119285992625879)
--(axis cs:100,0.109248886685838)
--(axis cs:50,0.105919747476573)
--(axis cs:20,0.102396440475025)
--cycle;

\addplot [very thick, orange, mark=*, mark size=3, mark options={solid}]
table {%
20 0.00152517914772034
50 0.00135141086578369
100 0.00140431380271912
200 0.00140112960338593
300 0.00139058621724447
};
\addlegendentry{\indepalg}
\addplot [very thick, blue, mark=*, mark size=3, mark options={solid}]
table {%
20 0.00260766983032227
50 0.00255541753768921
100 0.00254772424697876
200 0.00260705780982971
300 0.00250507966677348
};
\addlegendentry{\algknowngraph}
\addplot [very thick, green, mark=*, mark size=3, mark options={solid}]
table {%
20 0.0977407300472259
50 0.104423632621765
100 0.107207274675369
200 0.115530450701714
300 0.107914847056071
};
\addlegendentry{\algunknowngraph}
\end{axis}

\end{tikzpicture}}
        \caption{$10$ Products.}
        \label{fig:exp:runningtime:10prods}
    \end{subfigure}
    \hfill
    \begin{subfigure}[b]{0.325\textwidth}
        \resizebox{\linewidth}{!}{% This file was created with tikzplotlib v0.10.1.
\begin{tikzpicture}

\definecolor{darkgray176}{RGB}{176,176,176}
\definecolor{green}{RGB}{0,128,0}
\definecolor{lightgray204}{RGB}{204,204,204}
\definecolor{orange}{RGB}{255,165,0}

\begin{axis}[
height=7cm,
width=8cm,
legend cell align={left},
legend style={
  fill opacity=0.8,
  draw opacity=1,
  text opacity=1,
  at={(0.96,0.6)},
  anchor=east,
  draw=lightgray204
},
log basis y={10},
tick align=outside,
tick pos=left,
xlabel={Rounds},
ylabel={Time per Round [s]},
x grid style={darkgray176},
xmin=13, xmax=307,
xtick style={color=black},
y grid style={darkgray176},
ymin=0.0003, ymax=0.85,
ymode=log,
ytick style={color=black},
grid=both,
grid style={dashed,gray!50!white},
ytick={1e-05,0.0001,0.001,0.01,0.1,1},
yticklabels={
  \(\displaystyle {10^{-5}}\),
  \(\displaystyle {10^{-4}}\),
  \(\displaystyle {10^{-3}}\),
  \(\displaystyle {10^{-2}}\),
  \(\displaystyle {10^{-1}}\),
  \(\displaystyle {10^{0}}\)
}
]
\path [draw=orange, fill=orange, opacity=0.3]
(axis cs:20,0.00270924159517761)
--(axis cs:20,0.00247850111493592)
--(axis cs:50,0.00262539159159321)
--(axis cs:100,0.00264665096722974)
--(axis cs:200,0.00275177130812224)
--(axis cs:300,0.00275678535603737)
--(axis cs:300,0.00278750963704531)
--(axis cs:300,0.00278750963704531)
--(axis cs:200,0.00279425492173616)
--(axis cs:100,0.00278547650850879)
--(axis cs:50,0.00295491541524273)
--(axis cs:20,0.00270924159517761)
--cycle;

\path [draw=blue, fill=blue, opacity=0.3]
(axis cs:20,0.00578636416932552)
--(axis cs:20,0.00457428207853825)
--(axis cs:50,0.00494973756873016)
--(axis cs:100,0.00493501896836217)
--(axis cs:200,0.0052688930600429)
--(axis cs:300,0.00495764748475273)
--(axis cs:300,0.00500321213184476)
--(axis cs:300,0.00500321213184476)
--(axis cs:200,0.00686948174540754)
--(axis cs:100,0.0054942274572665)
--(axis cs:50,0.00573454187310333)
--(axis cs:20,0.00578636416932552)
--cycle;

\path [draw=green, fill=green, opacity=0.3]
(axis cs:20,0.386337763451223)
--(axis cs:20,0.352175088740702)
--(axis cs:50,0.42841029923639)
--(axis cs:100,0.440435235088741)
--(axis cs:200,0.461298141089182)
--(axis cs:300,0.497042801738861)
--(axis cs:300,0.505253550329722)
--(axis cs:300,0.505253550329722)
--(axis cs:200,0.478171453389425)
--(axis cs:100,0.449678828651036)
--(axis cs:50,0.471462891643432)
--(axis cs:20,0.386337763451223)
--cycle;

\addplot [very thick, orange, mark=*, mark size=3, mark options={solid}]
table {%
20 0.00259387135505676
50 0.00279015350341797
100 0.00271606373786926
200 0.0027730131149292
300 0.00277214749654134
};
\addlegendentry{\indepalg}
\addplot [very thick, blue, mark=*, mark size=3, mark options={solid}]
table {%
20 0.00518032312393188
50 0.00534213972091675
100 0.00521462321281433
200 0.00606918740272522
300 0.00498042980829875
};
\addlegendentry{\algknowngraph}
\addplot [very thick, green, mark=*, mark size=3, mark options={solid}]
table {%
20 0.369256426095962
50 0.449936595439911
100 0.445057031869888
200 0.469734797239304
300 0.501148176034292
};
\addlegendentry{\algunknowngraph}
\end{axis}

\end{tikzpicture}}
        \caption{$20$ Products.}
        \label{fig:exp:runningtime:20prods}
    \end{subfigure}
    \caption{Running time of the algorithms normalized by the number of rounds ($10$ trials, mean $\pm$ $95\%$ C.I.).}
    \label{fig:exp:runningtime}
\end{figure}

The algorithm proposed in this paper has been designed to be lightweight and to overcome several limitations induced by the adopted learning model (\ie GPs), which generally scale poorly with respect to the number of samples. As a result, all experiments can be efficiently executed on a standard consumer laptop. Specifically, all tests were conducted on a MacBook Pro equipped with an Apple Silicon M4 processor and $24$~GB of RAM. All experiments were run on a single core, although the algorithm can be partially parallelized. Figure~\ref{fig:exp:runningtime} reports the normalized running time of the algorithm for $5$ (Figure~\ref{fig:exp:runningtime:5prods}), $10$ (Figure~\ref{fig:exp:runningtime:10prods}), and $20$ (Figure~\ref{fig:exp:runningtime:20prods}) products for different target horizon $T\in\{20, 50, 100, 200, 300\}$, corresponding to the circle marks in the figures.
All plots share the same scale to facilitate comparison. Each figure shows the running time of a single execution, normalized by the number of rounds. We observe that, thanks to the heteroscedastic formulation introduced, the algorithm does not suffer from scalability issues with respect to the time horizon. A closer inspection of the results reveals that both \indepalg and \algknowngraph scale linearly with the number of products. Conversely, the version with an unknown graph structure, \algunknowngraph, is slightly more computationally demanding; however, it still exhibits a very reasonable running time. It is worth noting that the reported computational request of \algunknowngraph corresponds to a worst-case scenario, as here the integer optimization problem is solved at every step. In practice, this optimization can be performed less frequently, which would bring the computational performance of \algunknowngraph close to that of \algknowngraph.
\section{Related Works}
\label{sec:related}

In this section, we review the main contributions related to dynamic pricing, with a focus on studies employing machine learning approaches and on research concerning complementary products. For a comprehensive survey on dynamic pricing methodologies, readers are referred to the work of~\citet{den_boer_dynamic_2015}.

Among the available ML frameworks, \emph{multi-armed bandits} (MABs) have emerged as one of the most widely adopted techniques for dynamic pricing, as they effectively address the exploration-exploitation trade-off and ensure efficient sample utilization. The foundational study that introduced the concept of active learning of the demand curve in a pricing context was conducted by~\citet{rothschild_two-armed_1974}. Subsequent research expanded on this idea in various ways. \citet{kleinberg_value_2003} approached the challenge of continuous demand functions by discretizing price values and providing formal performance guarantees. Parametric versions of the demand function were later proposed by~\citet{besbes_dynamic_2009} and~\citet{broder_dynamic_2012}. \citet{besbes_surprising_2015} imposed monotonicity constraints on the demand model and demonstrated the effectiveness of linear models for demand estimation. Building on these ideas, \citet{shukla_dynamic_2019} developed a pricing algorithm that integrates customer-specific characteristics and assumes a monotonic willingness-to-pay function. \citet{mussi2023dynamic} propose a MAB algorithm to learn demand curves integrating a data-driven approach to volume discounts.

\subsection{Complementary Products}
Among studies that explicitly consider complementarities in pricing, \citep{mulhern_implicit_1991} is the earliest contribution empirically demonstrating that adopting multi-product pricing strategies can improve retail performance, although their focus was on product bundling rather than online learning. After that, several works tried to exploit complementary relations for pricing. \citet{sun_exploiting_2015} applied association rule mining to uncover implicit item-to-item relationships while accounting for their asymmetric nature. \citet{zhao_improving_2017} constructed a directed graph of complementary products using natural language processing and the Skip-Gram model~\citep{mikolov2013exploiting} to generate product embeddings. \citet{wang_path-constrained_2018} combined catalog metadata with transaction information to construct a graph capturing both substitutability and complementarity, applying category and path constraints to identify new connections. \citet{feng_substitutability_2018} modeled substitutability and complementarity within a discrete choice framework, but their approach lacks the adaptability and uncertainty modeling provided by ML techniques. \citet{stradi2024primaldualonlinelearningapproach} study the problem of dynamic pricing for complementary items shown sequentially to customers under uncertainty and sales constraints. It models the task as a constrained Markov decision process and develops a primal-dual online algorithm. Then, there are approaches, such as~\citep{mcauley_inferring_2015}, that rely heavily on textual data to infer complementarity, which may not accurately reflect the actual purchasing dynamics within a specific e-commerce environment, where transaction data provides a more accurate signal for pricing.
Additionally, most of these studies require extensive datasets, including textual descriptions, product reviews, or co-view information, whereas the present work assumes access only to transaction data to infer complementary relations. \citet{hao_p-companion_2020} developed a graph-based representation learning model that integrates textual and behavioral data to recommend diverse, complementary items, while~\citet{xu_product_2020} leveraged co-view data and product descriptions to build knowledge graph embeddings. Although these methods effectively capture relationships between products, their primary goal differs from the focus of this paper. In particular, they do not aim to simplify the graph structure or jointly optimize pricing decisions to exploit relations that generate a higher revenue/profit. On the theoretical perspective, \citet{mussi2025factored} demonstrate that synchronization among actions is crucial for achieving optimal performance in sequential decision-making problems.
\section{Discussion and Conclusions}
\label{sec:conclusions}

In this paper, we addressed the problem of finding the optimal pricing strategy for products exhibiting complementary relations. We began by formalizing the problem and defining the associated learning objective. Then, we introduced \algname, a novel three-stage algorithm designed for online learning in such environments. The algorithm first aggregates substitutable products using the available information. Subsequently, it identifies complementary relations through a specifically crafted integer programming formulation and estimates the corresponding demand curves using a statistically and computationally efficient implementation of Gaussian Processes that leverages heteroscedasticity to reduce computational complexity. To the best of our knowledge, this is the first complete pipeline that jointly identifies complementary products and optimizes pricing decisions while explicitly accounting for computational constraints. We conducted an extensive experimental campaign to validate the soundness and effectiveness of our approach. The results show that knowing the structure of complementarities significantly simplifies the learning process, leading to faster convergence and more stable performance. When such relations are unknown, estimating them through our proposed approach provides a clear improvement over treating products as independent. However, some improvement can still be beneficial to enhance the stability of the integer programming stage and mitigate the effects of over-exploration. Overall, the experiments confirm the robustness of the proposed framework and its ability to adapt to different levels of product interaction.

\paragraph{Future Works.}~~~In the short term, we plan to analyze the computational complexity of the integer programming component presented in Section~\ref{sec:algorithm:compl}, and to design heuristic or approximate solutions improving stability and scalability. In the longer term, we aim to integrate a model of customer rationality into the framework, enabling a more realistic representation of consumer behavior and a richer set of strategic interactions in dynamic pricing environments.

\bibliographystyle{unsrtnat}
\bibliography{biblio}

\end{document}